\documentclass{bmvc2k}

\title{Camera Elevation Estimation from a Single Mountain Landscape Photograph}

\addauthor{Martin \v{C}ad\'{i}k}{cadik@fit.vutbr.cz}{1}
\addauthor{Jan Va\v{s}\'{i}\v{c}ek}{xvasic21@stud.fit.vutbr.cz}{1}
\addauthor{Michal Hradi\v{s}}{ihradis@fit.vutbr.cz}{1}
\addauthor{Filip Radenovi{\' c}}{filip.radenovic@cmp.felk.cvut.cz}{2}
\addauthor{Ond{\v r}ej Chum}{chum@cmp.felk.cvut.cz}{2}

\addinstitution{
 CPhoto@FIT, \bmvaUrl{http://cphoto.fit.vutbr.cz},\\
 Faculty of Information Technology,\\
 Brno University of Technology,\\
 Brno, Czech Republic
 \vspace{3mm}}
\addinstitution{
 Center for Machine Perception,\\
 Department of Cybernetics,\\
 Faculty of Electrical Engineering,\\
 Czech Technical University in Prague,\\
 Prague, Czech Republic}

\runninghead{\v{C}ad\'{i}k \etal}{Camera Elevation Estimation from a Single Photograph}

\usepackage{xspace}

\def\etal{\emph{et al}\bmvaOneDot}
\newcommand{\ie}{\mbox{\emph{i.e.\xspace}}} 

\usepackage{color}
\usepackage[percent]{overpic}

\renewcommand{\paragraph}[1]{\vspace{.3\baselineskip}\noindent{\bf{#1}}\ }

\begin{document}

\maketitle

\begin{abstract}
This work addresses the problem of camera elevation estimation from a single photograph in an outdoor environment. We introduce a new benchmark dataset of one-hundred thousand images with annotated camera elevation called Alps100K. We propose and experimentally evaluate two automatic data-driven approaches to camera elevation estimation: one based on convolutional neural networks, the other on local features. To compare the proposed methods to human performance, an experiment with 100 subjects is conducted. The experimental results show that both proposed approaches outperform humans and that the best result is achieved by their combination.

\end{abstract}

\section{Introduction}
\label{sec:intro}

In outdoor environments one of the more important and informative attributes is the elevation: the height of a geographic location above the sea level. Estimation of elevation has a long history~\cite{cajori29history}. Nowadays, elevation data are important for a number of applications, including earth sciences, global climate change research, hydrology, and outdoor navigation. 
%
%
Traditionally the assessment of elevation was the domain of geodesy, which offered several means to measure altitude. Among the most popular methods are barometric altimeters, 
trigonometric or leveling measurements, 
and Global Positioning System (GPS). 

\begin{figure}[t]\centering 
	\centering
        \begin{overpic}[width=0.18\linewidth]{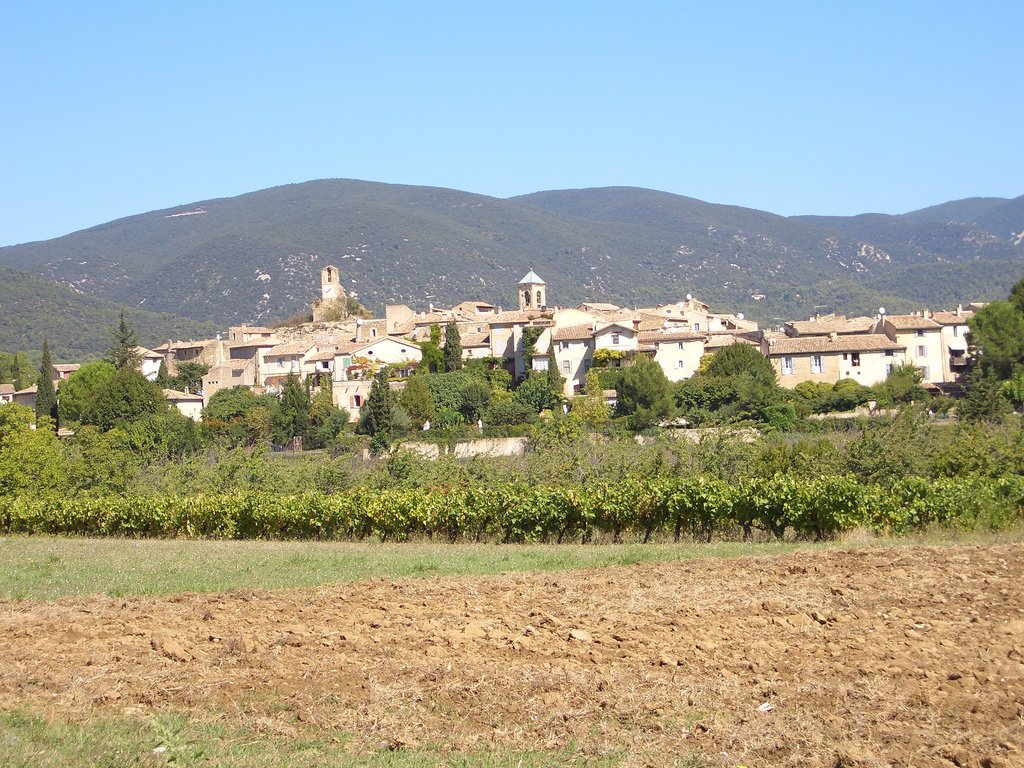}
        \put (1,60) {\footnotesize 217.5m}
        \put (1,52) {\tiny 43.7644N, 5.3622E}
        \end{overpic}
        \begin{overpic}[width=0.18\linewidth]{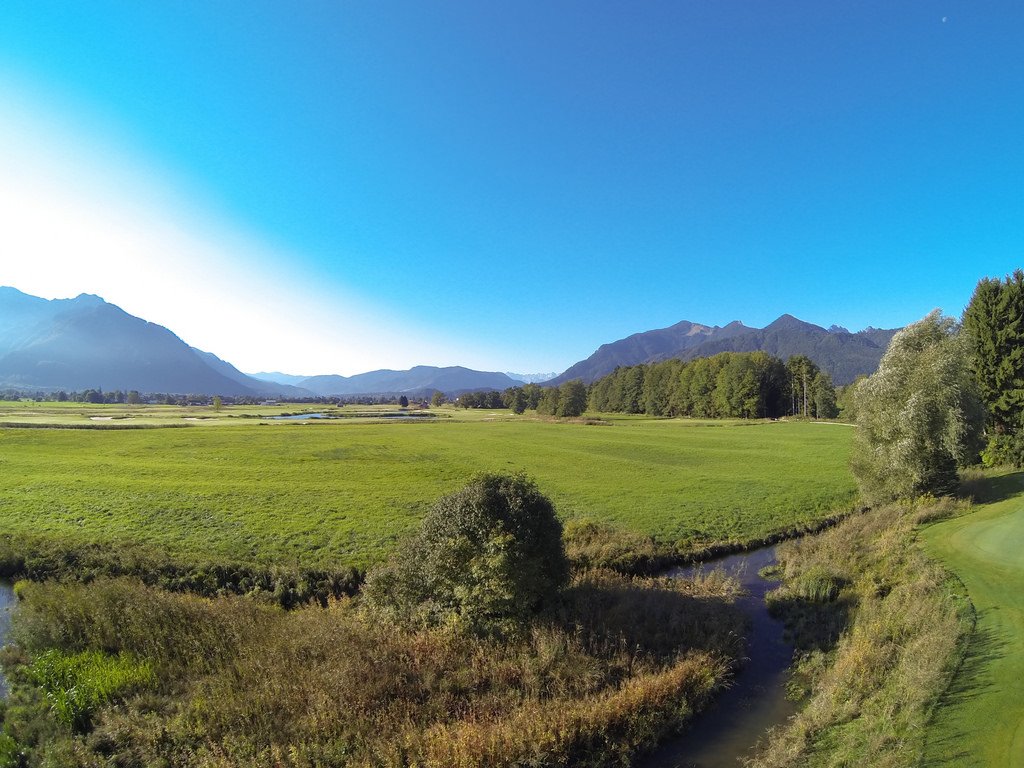}
        \put (1,60) {\footnotesize 529m}
        \put (1,52) {\tiny 47.7923N, 12.4608E}
        \end{overpic}
        \begin{overpic}[width=0.18\linewidth]{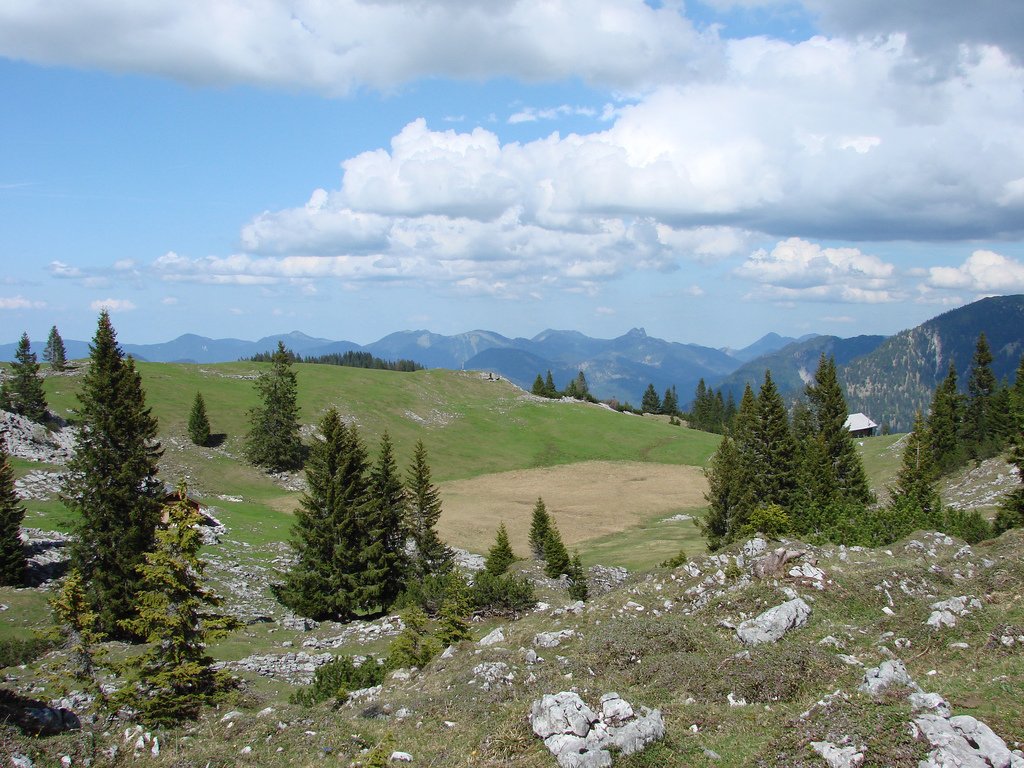}
        \put (1,60) {\footnotesize 1479m}
        \put (1,52) {\tiny 47.5178N, 11.4628E}
        \end{overpic}
        \begin{overpic}[width=0.18\linewidth]{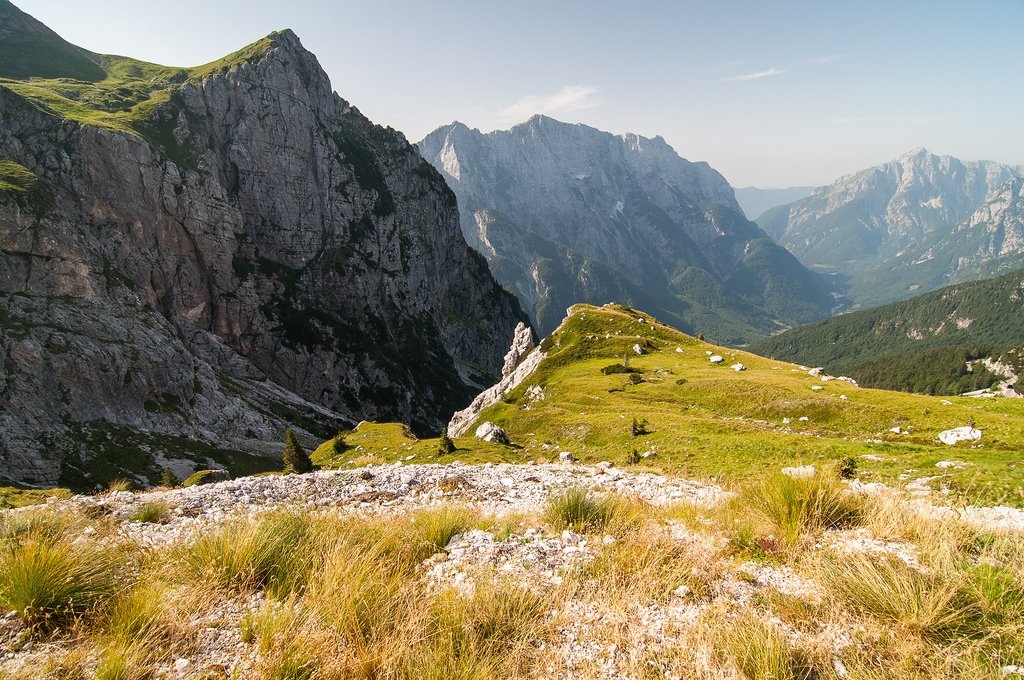}
        \put (1,60) {\footnotesize 1798m}
        \put (1,52) {\tiny 46.4384N, 13.6351E}
        \end{overpic}
        \begin{overpic}[width=0.18\linewidth]{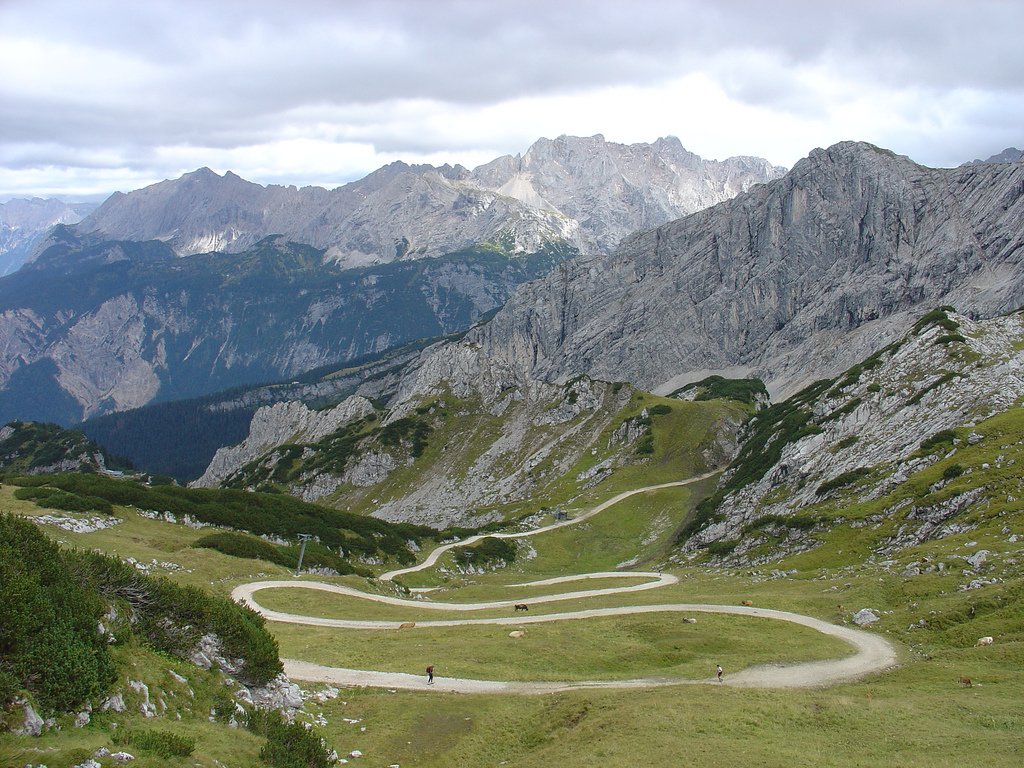}
        \put (1,60) {\footnotesize 2395m}
        \put (1,52) {\tiny 47.4274N, 11.044E}
        \end{overpic}
        \begin{overpic}[width=0.18\linewidth]{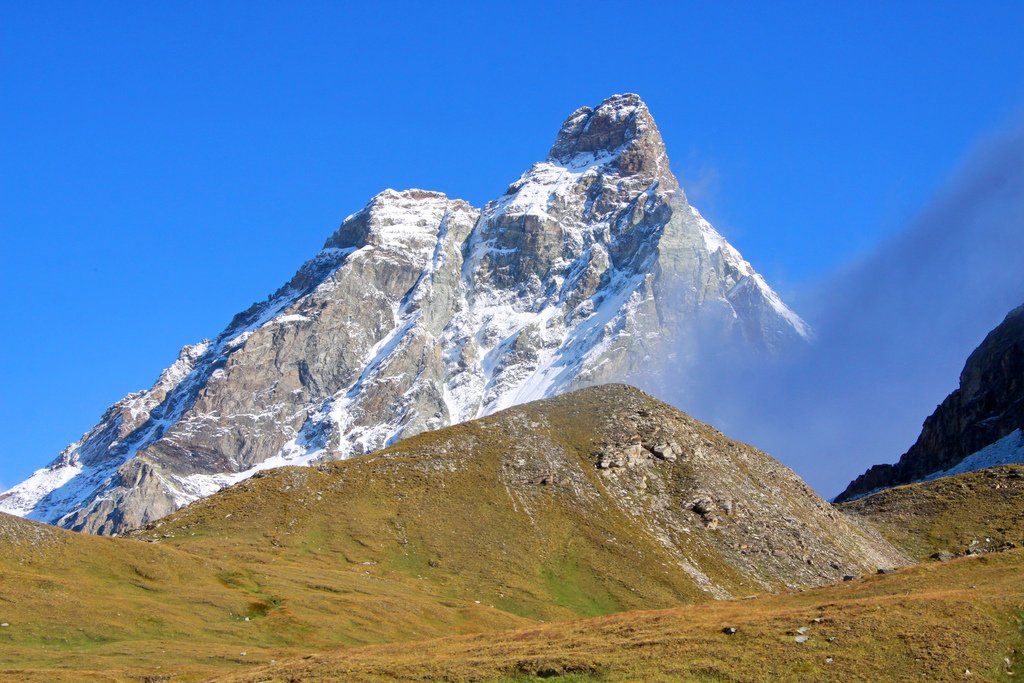}
        \put (1,60) {\footnotesize 2504.5m}
        \put (1,52) {\tiny 45.9449N, 7.6527E}
        \end{overpic}
        \begin{overpic}[width=0.18\linewidth]{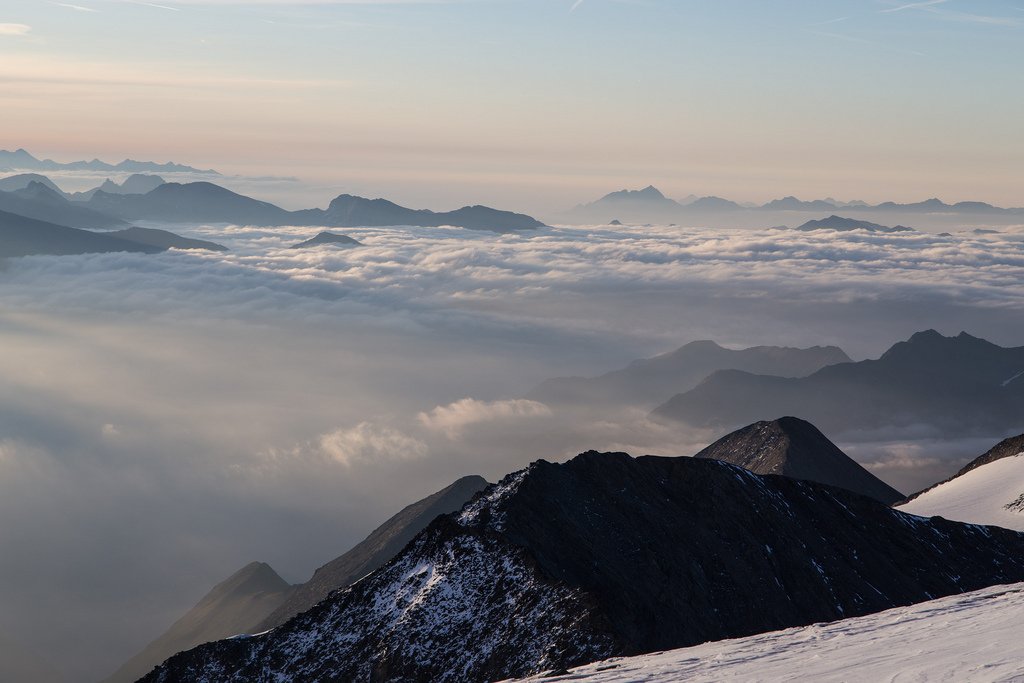}
        \put (1,60) {\footnotesize 3375.5m}
        \put (1,52) {\tiny 47.0709N, 12.7011E}
        \end{overpic}
        \begin{overpic}[width=0.18\linewidth]{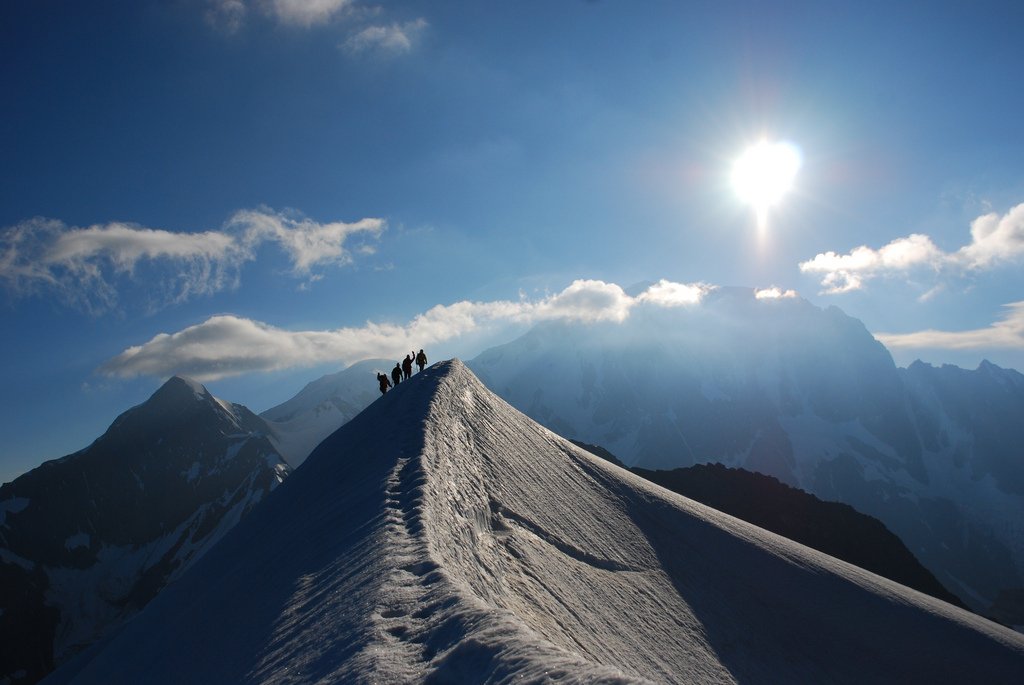}
        \put (1,60) {\footnotesize 3503m}
        \put (1,52) {\tiny 45.8113N, 6.7909E}
        \end{overpic}
        \begin{overpic}[width=0.18\linewidth]{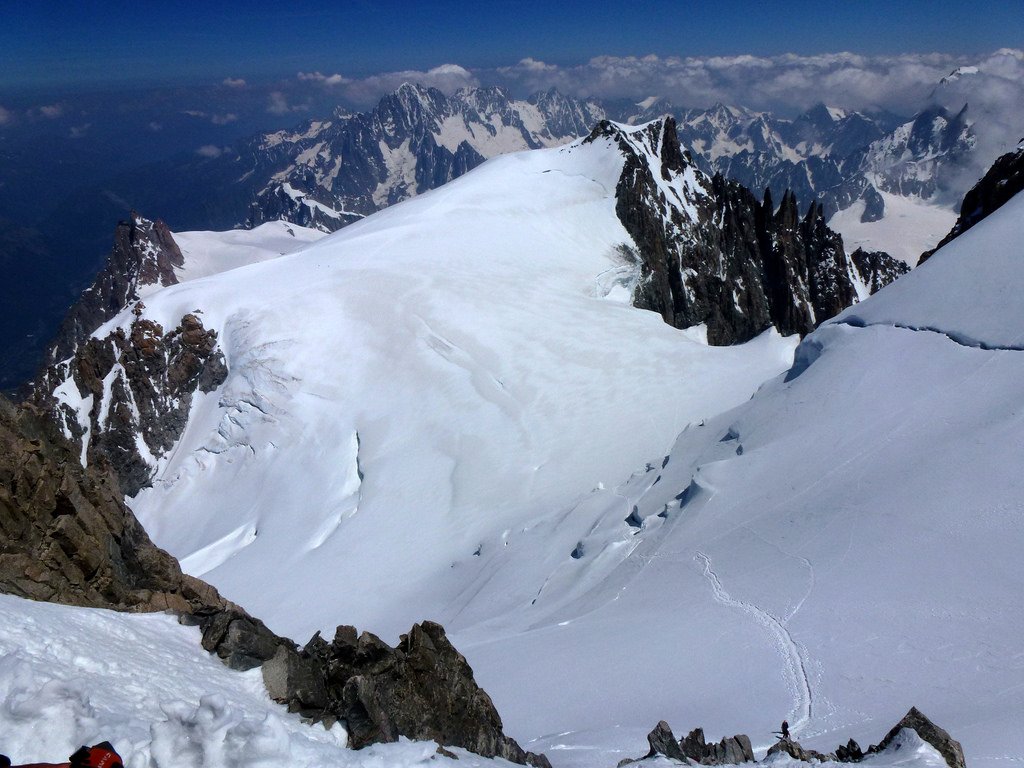}
        \put (1,60) {\footnotesize 4158.5m}
        \put (1,52) {\tiny 45.8463N, 6.8737E}
        \end{overpic}
        \begin{overpic}[width=0.18\linewidth]{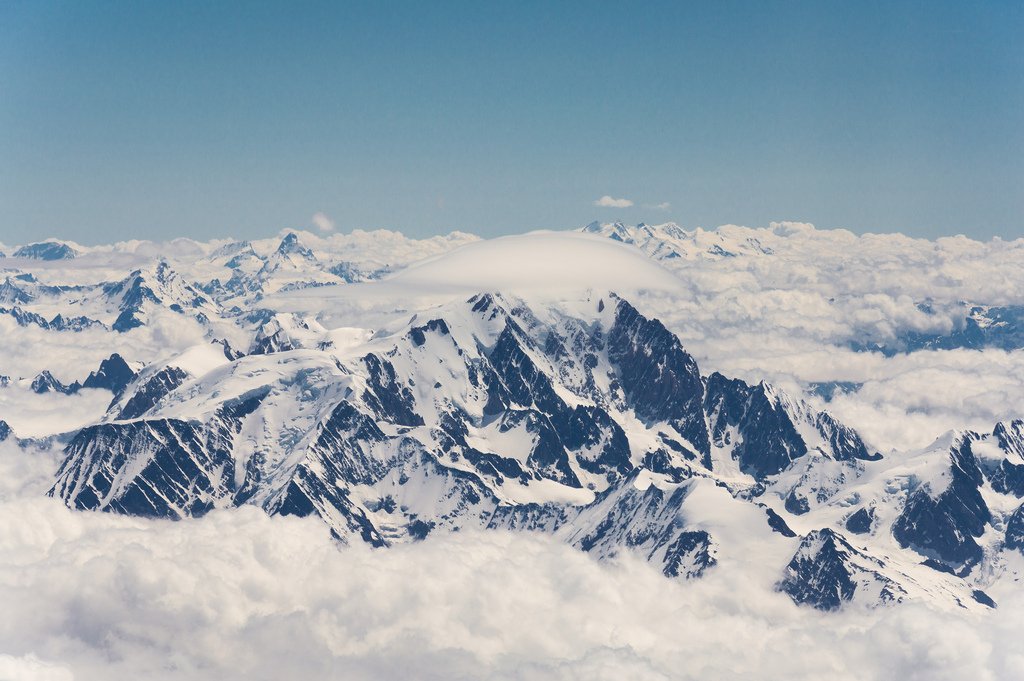}
        \put (1,60) {\footnotesize 4782m}
        \put (1,52) {\tiny 45.8336N, 6.8616E}
        \end{overpic}
	\caption[]{A sample from the new benchmark dataset Alps100K~\cite{cadik2015supplementary}. Image credits - flickr users: Allie$\_$Caulfield, Golf Resort Achental Team, Erik, Guillaume Baviere, Tadas Bal\v{c}iunas, antoine.pardigon, twiga269, Karim von Orelli.}
	\label{fig:dataset}
\end{figure}
\begin{figure}[t]
	\centering
	\begin{minipage}{0.43\textwidth}
	\includegraphics[width=\linewidth]{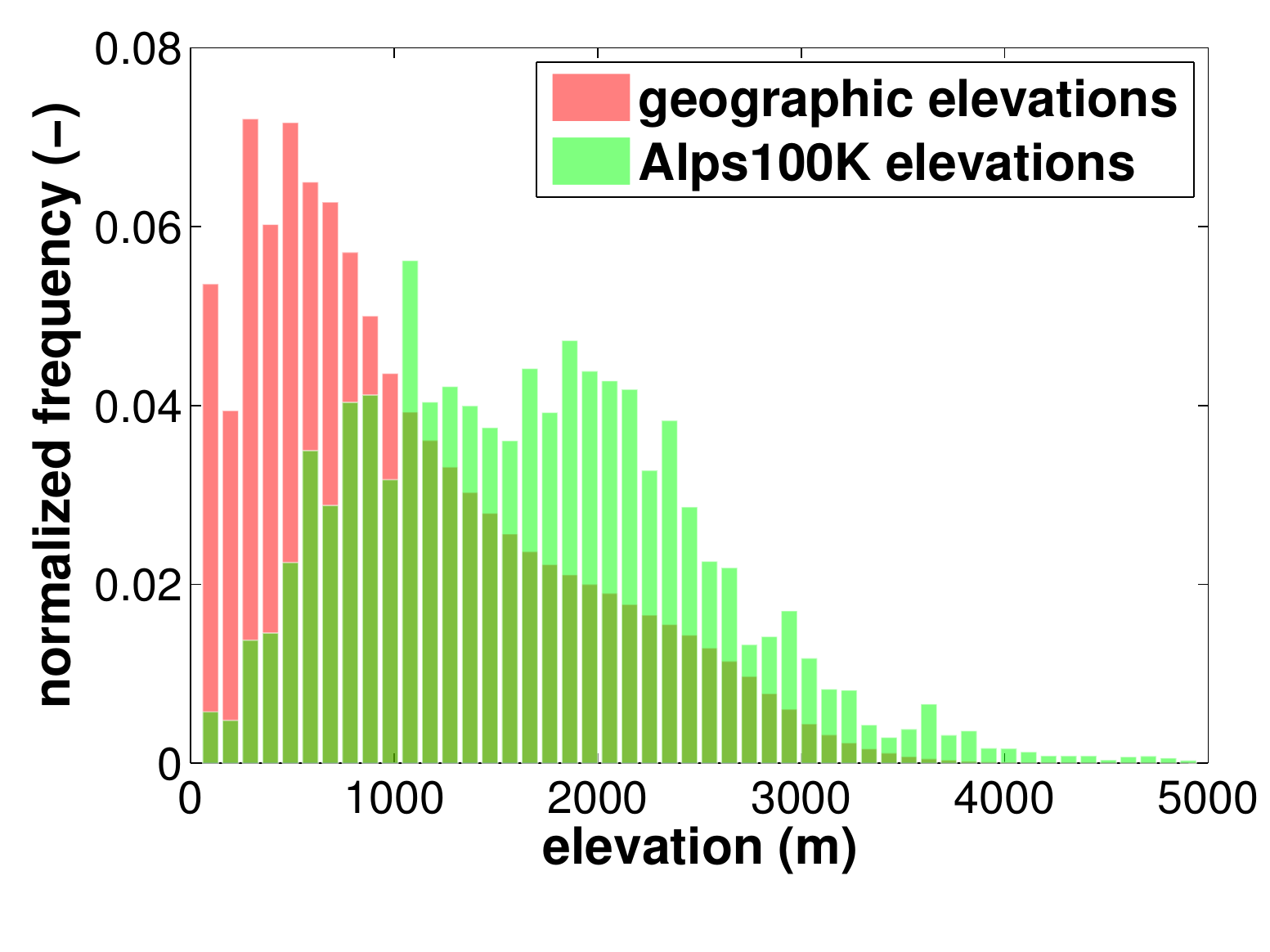}
	\end{minipage}
	\hspace{2mm}
	\begin{minipage}{0.5\textwidth}
	\vspace{-4mm}
	\includegraphics[width=0.88\linewidth]{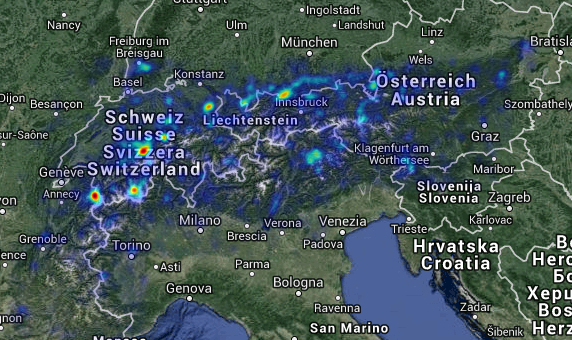}
	\includegraphics[width=0.08\linewidth]{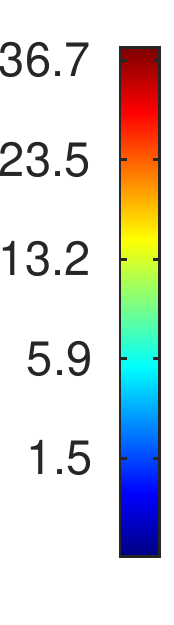}
	\end{minipage}
\caption{ Left: normalized elevation histogram of the Alps mountain range (red) and the distribution of elevations in the Alps100K dataset (green). Right: geographic coverage of the new dataset. 
}
	\label{fig:dataset_histogram}
\end{figure}

A rich natural heritage is captured and expanded everyday in the form of landscape photos and videos with an ever-growing geographic coverage of outdoor areas, see Figs.~\ref{fig:dataset} and~\ref{fig:dataset_histogram}.
Unfortunately, almost all currently available photos and videos lack elevation information. Moreover, a majority of them do not even contain the GPS coordinates. In this paper, we tackle the problem of estimating the elevation from visual data only. Automatic annotation of images with an accurate estimate of the elevation can be exploited in a number of applications ranging from leisure activities and tourism applications to
image database exploration, education purposes and geographic games.



\paragraph{Contributions.}
In order to evaluate elevation estimation methods, we introduce a new dataset containing approximately 100K images of natural environments accompanied with the GPS and elevation information (Section~\ref{sec:dataset}). 
We propose two methods of elevation estimation from image content: one based on convolutional neural networks (CNN)~\cite{zhou2014places}, the other exploiting bag-of-words (BOW) based image retrieval~\cite{Sivic-ICCV03,Jegou-ECCV12}.
The proposed methods are compared to a human performance in Section~\ref{sec:results}. To estimate the human performance on this task, an experiment was conducted counting 100 subjects (Section \ref{sec:human}). The proposed automatic elevation estimation methods outperform knowledgeable humans, and, moreover, the hybrid combination of BOW with CNNs results in the best predictions.


\section{Related work}
\label{sec:related_work}
We are not aware of any attempt to predict the camera elevation from visual information contained in a landscape photograph of a natural environment. However, the related field of research is the one of \emph{visual geo-localization}. If we were able to localize the position of the camera, the task of elevation estimation would reduce to a simple query into a geo-referenced terrain model. Unfortunately existing geo-localization methods~\cite{Hays2008,Baboud2011,BSKP12,zamir14image,tzeng13user} are neither robust nor sufficiently accurate for this task. Conversely, the methods presented in this paper may improve geo-localization techniques by reducing the search space to only probable camera elevations. 

Height above the ground is among the most important information for navigation of \emph{Unmanned Aerial Vehicles} (UAV). The problem of elevation estimation for UAVs has been attacked using computer vision because it has several advantages: it is a passive system, has low energy consumption, and visual information can be reused for navigation or localization. The proposed solutions are based on artificial ground-located landmarks~\cite{saripalli02vision,Garcia-Pardo-2001-85}, optical flow~\cite{beyeler06vision,srinivasan03}, stereoscopic vision~\cite{jung03high,meingast04vision,eynard10uav}, 
and machine learning~\cite{cherian09autonomous}. 
Most of these methods assume that the camera's viewpoint is oriented to the ground and that the camera parameters are known. In contrast, in this work we aim to assess the \emph{absolute} elevation of the camera. Moreover, the hereby proposed methods work on ordinary photographs of natural outdoor environments. Our scenario is more challenging, since both camera orientation and calibration information is missing. 

The lack of work devoted to elevation estimation may be explained by the absence of a suitable dataset. The recently published Places 205~\cite{zhou2014places} is a dataset for training scene classifiers. It contains 205 scene categories and almost 2.5 million images of which a subset could be selected for our task. Unfortunately, Places 205 contains neither elevation information nor GPS coordinates. IM2GPS dataset~\cite{Hays2008} does contain GPS coordinates for each image; however, these images are mostly captured in cities with a significant bias towards landmarks like the Eiffel tower or the Sydney opera house. This renders IM2GPS dataset useless for our purpose because we focus on landscape photos of outdoor environments, like the ones shown in Fig.~\ref{fig:dataset}.

\section{Alps100K: a new dataset}
\label{sec:dataset}

We introduce a new dataset of almost 100K annotated (GPS coordinates, elevation, EXIF if available) outdoor images from mountain environments. The collection covers vast geographic area of the Alps, the highest range in Europe; therefore we name it Alps100K. 
The images exhibit high variation in elevation as well as in landscape appearance. Furthermore, the collection spans all the seasons of the year. 
To the best of our knowledge, this is the first dataset of this kind.  
It contains test sets to evaluate elevation estimation performance (see Section~\ref{sec:human} for human performance and Section~\ref{sec:results} for results of the proposed automated methods). A large proportion of the dataset serves as a training set for the data-driven approaches.


\paragraph{Dataset acquisition.} First we create a list of all hills and mountain peaks located in the seven Alpine countries (Austria, France, Germany, Italy, Liechtenstein, Slovenia, Switzerland) from the OpenStreetMap database~\cite{Curran2012}. The list of hill names is used to query the Flickr\footnote{\url{http://www.flickr.com}} photo hosting service. In order to increase the ratio of outdoor images certain tags, such as wedding, family, indoor, still life, are excluded. Only images containing information about the camera location are kept. Out of 1.2M crawled images, about 400K are unique and inside the Alps region.

To cull irrelevant (non-landscape) images 
a state-of-the-art scene classifier~\cite{zhou2014places} is applied.
%
Probabilities of 205 scene categories are assigned to each image. 
We experimentally select 28 categories (mountain, valley, snowfield, etc.) and keep only those images whose cumulative probability in those categories exceeds 0.5. This step significantly improves the relevance of the dataset at the expense of reducing the number of images to circa 25\%.  

Finally the elevation of the camera is inferred from the GPS coordinates via the digital elevation model\footnote{Available from \url{http://www.viewfinderpanoramas.org}}. This model covers the Alps with 24 meter spaced samples.
The collection contains \emph{98136 outdoor images} that span almost all possible elevations observed in the Alps [0, 4782m]. Fig.~\ref{fig:dataset_histogram} left compares the elevation distribution of the Alps surface and the elevations in the dataset, Fig.~\ref{fig:dataset_histogram} right shows the spatial distribution of the collected images. Geographically the dataset covers virtually all the regions of the Alps with obvious concentrations in tourist spots (e.g., around Zermatt village in Switzerland). 
The EXIF information is available for 41364 images, which is 42\% of the Alps100K dataset. 
The the presented dataset along with elevation and GPS meta-data is available at the project webpage~\cite{cadik2015supplementary}. 



\section{Human performance in elevation estimation task}
\label{sec:human}
In this section we measure the ability of humans to estimate camera elevation from an image. The achieved accuracy is subsequently compared to results achieved by the methods proposed in this paper.


During the experiment 100 participants were asked to estimate the {\em camera} (not the depicted scene) elevation for each of the 50 test images.
We utilized a custom web-based interface where the participants assessed the elevation of each test image (for an example see Fig.~\ref{fig:dataset}) using a slider (see more details on the experiment in the supplementary material available at the project webpage~\cite{cadik2015supplementary}). The elevation of the test images ranged in [79m, 4463m] (see green crosses in Fig.~\ref{fig:boxplot}). The images were presented in randomized order, at the resolution of 750$\times$500px.  
%
%
After the experiment was finished participants filled out a questionnaire where additional information about their age, experience with the Alps, highest reached elevation, etc., was gathered. The subjects needed 10 minutes on average to complete the experiment.  

\subsection{Experimental results}


\paragraph{Are humans able to estimate the elevation from an image?} We use the analysis of variance (ANOVA) test~\cite{Bewick2004} which determines whether there is a systematic effect of an image (\ie, the elevation) in human elevation predictions, or whether the predictions are random. Formally, we state the null hypothesis $H_0$ as follows: there is no significant difference between elevation predictions for the test images. This hypothesis is clearly rejected ($F(49,4950)=165.09,p<0.001$), meaning that humans \emph{can} indeed estimate the camera elevation from the visual information in images. 

\begin{figure*}[t]
	\centering 
	\rotatebox{90}{\mbox{}\hspace{3cm}\footnotesize{elevation}}\includegraphics*[width=.9\linewidth, trim=21mm 13mm 20mm 7mm]{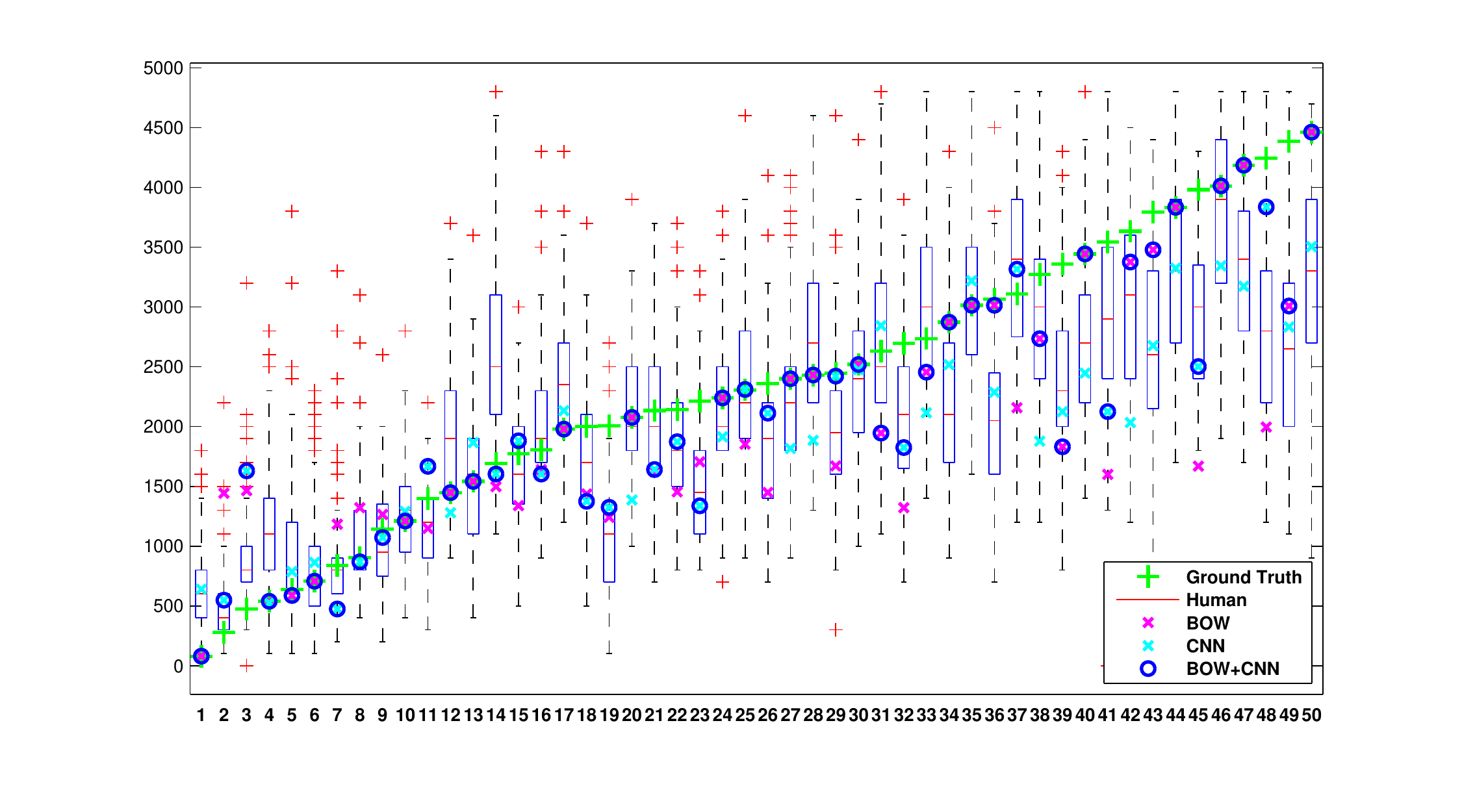}\\
    \footnotesize{index to the 50 test images}
    	\caption{ Comparison of the elevation estimation by humans and the proposed methods (CNN, BOW and combination).
Blue boxes show the span of the human predictions: the red mark is the median, the edges of the box are the $25^{\mathrm {th}}$ and $75^{\mathrm {th}}$
percentiles respectively, the whiskers extend to extreme human guesses that are not considered outliers, and outliers are plotted individually as '\textcolor{red}{+}'.
}
	\label{fig:boxplot}
\end{figure*}

\paragraph{How well can humans predict the camera elevation?} 
The predictions for each test image along with the ground truths are plotted in Fig.~\ref{fig:boxplot}.
The overall root-mean-squared error (RMSE) of human elevation predictions is $\mathrm{RMSE}(H)=879.95\mathrm{m}.$ 
It can be observed that people underestimate high elevations, \ie, elevations above 3000m. In accordance with this, variance of the elevation predictions grows for high altitudes as well. 


The effect of human age, sex, experience and other factors collected in post-experiment questionnaire was also analyzed. None of those factors were found statistically significant. However, it is worth noting that our participants were either living in the Alps, had exceptional experience with outdoor sports, or both. Accordingly, the reported average prediction errors should be taken as rather conservative ones.

\section{Automatic elevation estimation from landscape photo}
\label{sec:automatic}

In this section we propose two approaches to estimate elevation from the visual content. We use popular convolutional neural networks and methods based on local features and combine them to exploit each approach's strengths in a single hybrid method. 

\subsection{Convolutional neural networks (CNN)}
\label{sec:CNN}

The elevation estimation task can be treated as a standard regression problem where the goal is to directly predict the real-valued elevation given the pixel data of a single photo. Convolutional neural networks have proven to be the state-of-the-art in various image-based machine learning tasks including object and scene classification~\cite{zhou2014places}, object detection~\cite{girshick14CVPR}, semantic segmentation~\cite{Long2014}, and facial recognition~\cite{Taigman2014}. We build upon the previous successes and apply large convolutional networks to the Alps100K dataset. Considering the relatively small size of the dataset, we initialize the networks from a network previously trained on the Places205 dataset~\cite{zhou2014places}, which includes a rich collection of outdoor scenes among its 2.5 million images. Additionally, we extend the  network inputs by EXIF data, which carries information indicative of possible weather conditions and camera settings.

\paragraph{Convolutional network architecture.}
The basic network architecture follows previous successful work on image classification. Specifically, it is the same as the one used in the Caffe reference network\footnote{Available from the Caffe package  \url{http://caffe.berkeleyvision.org/model\_zoo.html}} and the Places-CNN network~\cite{zhou2014places}, which are in turn very similar to the network used by Krizhevsky et al.~\cite{Krizhevsky2012} to win the ImageNet challenge in 2012.

The main building blocks of the network (see Tab.~\ref{tab:CNN_architecture}) are convolutions followed by Rectified Linear Units (ReLU). First, second, and fifth convolutional layers are followed by max-pooling, each reducing resolution by a factor of two. The activations of the first and second convolutional layers are locally normalized~\cite{Krizhevsky2012}. The output of the convolutional part of the network is fed into a fully connected layer (fc6) with 4096 units. Weights of this part of the network are initialized from network Places-CNN\footnote{Available from \url{http://places.csail.mit.edu}}~\cite{zhou2014places}.

The final two layers of the network are fully connected and contain 2048 and 1 neurons, respectively. Weights of these layers were initialized from a normal distribution with standard deviation 0.005 and 0.02, respectively. The final activation function is linear and the optimization objective is Mean Squared Error (MSE). The network was trained by mini-batch Stochastic Gradient Descent with momentum.

\begin{table}[t]
\centering%
  \caption{Architecture of the convolutional network. Layers up to fc6 were initialized from Places-CNN network~\cite{zhou2014places}. Resolution of the input is $227 \times 227$ pixels.}
  \setlength{\tabcolsep}{2.5pt}
  \begin{tabular}{c|ccccccccccc}
	\hline
	\hline
	Layer       & conv1 & pool1 & conv2 & pool2 & conv3 & conv4 & conv5 & pool5 & fc6 & fc7 & fc8\\
	\hline
    units       & 96    & 96    & 256   & 256   & 384   & 384   & 256   & 256   & 4096  & 2048 & 1 \\
	kernel&$11 {\times} 11$&$3 {\times} 3$&$5 {\times} 5$&$3 {\times} 3$&$3 {\times} 3$&$3 {\times} 3$&$3 {\times} 3$&$3 {\times} 3$& - & - & -\\
    features    &$55 {\times} 55$&$27 {\times} 27$&$27 {\times} 27$&$13 {\times} 13$&$13 {\times} 13$&$13 {\times} 13$&$13 {\times} 13$&$6 {\times} 6$& - & - & -\\
    
   \end{tabular}
  \label{tab:CNN_architecture}
\end{table}

\subsection{Local features}
\label{sec:bow}

An alternative approach is based on the $k$-NN classifier~\cite{Duda-Pattern12}. Two efficient methods of obtaining nearest neighbours are considered: sparse high-dimensional bag-of-words (BOW) based image retrieval~\cite{Sivic-ICCV03}, and image retrieval with compact image representation~\cite{Jegou-ECCV12,Radenovic-ICMR15}. In both cases, the number of neighbours $k$ used to estimate the elevation is a function of the confidence in the retrieved nearest neighbour.

The BOW approach has been shown to perform well for specific object and place recognition, especially when combined with a spatial verification step~\cite{Philbin-CVPR07}, while the short-vector image representations~\cite{Jegou-ECCV12,Radenovic-ICMR15} obtained by a joint dimensionality reduction from multiple vocabularies show certain level of generalization power.

\paragraph{High-dimensional image representation (BOW).} The majority of image retrieval methods based on BOW representation follow the same procedure as introduced in~\cite{Sivic-ICCV03}.
%
%
First, local features~\cite{Mikolajczyk-IJCV04} such as multi-scale Hessian-Affine~\cite{Perdoch-CVPR09} are detected and described by an invariant $d$-dimensional descriptor such as SIFT~\cite{Lowe-IJCV04} or RootSIFT~\cite{Arandjelovic-CVPR12} for all images in the dataset.
%
Then the descriptor space is vector-quantized into a visual vocabulary: a $K$-means algorithm is performed on an independent training dataset to create $K$ clusters representing the visual words. In our paper, $K{=}1$M visual words was used. 
%
Finally, a histogram of occurrences of visual words is generated for each image followed by the inverse document frequency weighting (\textit{idf}), and sparse BOW vectors are obtained with dimensionality $D{=}K$.
%
%

To estimate the elevation of a photograph the photograph is used to query an elevation-annotated image database (the training part of Alps100K). Efficient retrieval via the inverted file structure~\cite{Sivic-ICCV03} is followed by a spatial verification step~\cite{Philbin-CVPR07} to re-rank the results. If the top ranked image is likely to be from the same location as the query image, \ie, the top ranked image is spatially verified with high confidence (more than $t_{sp}$ features pass the verification test), we use its elevation as the estimate. Otherwise, we use median of elevations of all retrieved $k$-NN images.

\paragraph{Short-vector image representation (mVocab).} Large scale image retrieval with short vectors has recently became popular as a method for reducing high computational or memory costs. In~\cite{Jegou-ECCV12,Radenovic-ICMR15} concatenated vocabularies of different origins followed by a joint dimensionality reduction are used as short image descriptors.


In our experiments we follow~\cite{Radenovic-ICMR15} and combine eight different visual vocabularies of 8K visual words each. The vocabularies are constructed over two different measurement regions and four power-law normalizations of SIFT descriptors. The region sizes are $r{\times}s$ and $1.5{\times}r{\times}s$ ($s$ is the scale of the detected feature, and $r{=}3\sqrt{3}$ is the standard relative scale change between detected region and the measurement region, as in~\cite{Perdoch-CVPR09}). The power-law normalization of SIFT descriptors ranges $\beta{=}0.4{,}0.5{,}0.6{,}1$, where $\beta{=}1$ is the original SIFT descriptors, while $\beta=0.5$ corresponds to RootSIFT~\cite{Arandjelovic-CVPR12}.

Camera elevation of landscape photographs is estimated by finding $k$-NN images from the training dataset and taking the weighted average of their elevations. Weights $\mathbf{w}$ are calculated using image dissimilarities $\mathbf{d}$, $\mathbf{w} {=} \max [ 0 , 1 {-} \mathbf{d} / ( w_t d_1 ) ]$, where $d_1$ is the dissimilarity of the top ranked image and $w_t$ is a constant, $w_t{\geq}1$. The number of retrieved images is fixed, but the number used from that set depends on their similarity to the top ranked image; in addition, we can control the number used from the retrieved set by the appropriate choice of parameter $w_t$. Specifically, only images that have dissimilarity up to $w_t d_1$ are used to compute weighted average of elevations.


\paragraph{Parameter choice.} We use the training dataset to learn the necessary parameters. For all experiments presented in this paper $t_{sp}{=}8$, $w_t{=}1.4$ and $k{=}100$. Slight changes do not significantly influence the presented results.

\subsection{Combining the elevation estimators} 
\label{sec:hybrid}


The BOW representation with spatial verification has been shown to perform well in a specific object or location recognition. Since people take and share photographs from similar locations, a natural approach is to recognize the specific location. On the other hand, BOW does not generalize meaning that it does not perform well on unseen scenes. Therefore, we propose a hybrid method that first tries to estimate the elevation by recognizing the location, and if that fails, \ie, no spatially verified image is retrieved, then by a secondary estimator: either mVocab or CNN.

\section{Results}
\label{sec:results}

In this section we experimentally evaluate the proposed methods. A \emph{test set} of 13148 images (13\% of the datased) is randomly selected from Alps100K. The rest of the dataset (\ie, 84988 images) is used for \emph{training}. The selected measure of performance is an overall root-mean-square error (RMSE) of elevation predictions with regards to the known ground truth elevations. The outcome is summarized in Tab.~\ref{tab:results}, where \emph{Baseline} denotes a simple elevation predictor that reports the mean elevation of the training dataset for all queries. For completeness we show the best achievable baseline RMSE values as well, \ie, using the mean of the \emph{testing} elevations (in italics).  

\begin{figure}[ht]
	\centering
	\includegraphics[width=0.45\linewidth]{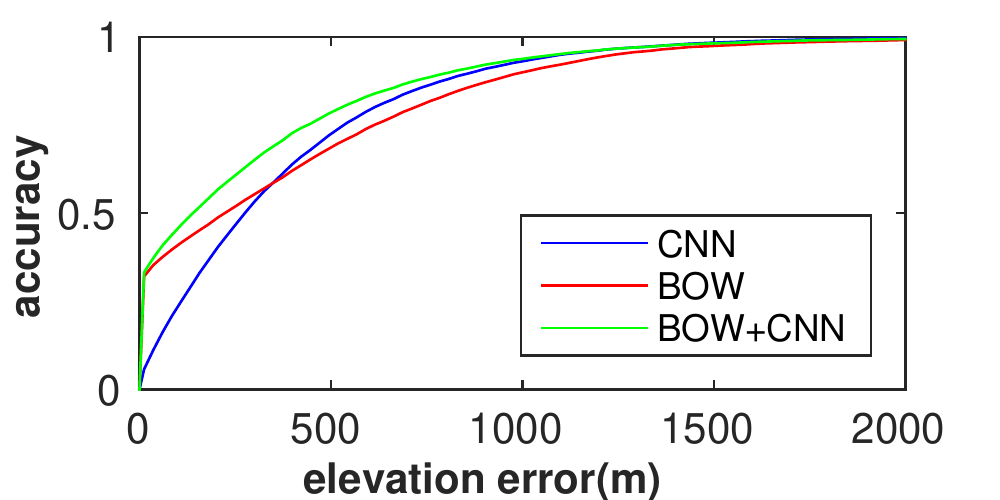}
    \hspace{0.08\linewidth}
	\includegraphics[width=0.45\linewidth]{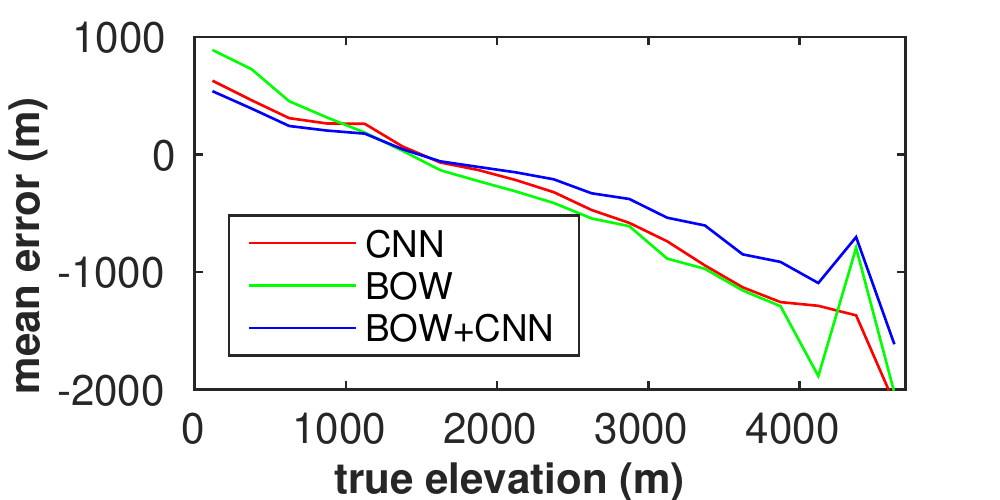}
	\caption{Left: cumulative elevation prediction accuracy. Right: dependence of prediction bias on image elevation. }
	\label{fig:cumul_graph}
\end{figure}

All of the proposed methods perform better than the baseline. The best prediction accuracy among the individual methods is achieved by neural networks (CNN); however, this is significantly improved by combining BOW with CNN. For a better understanding of these results, we plot the fraction of correct predictions as a function of the elevation error, see~Fig.~\ref{fig:cumul_graph}, left. 
For more than one third of the test images, the BOW approach spatially verifies the query image, which results in extremely low error rate. The specific place recognition performed by the BOW approach seems to work well for this particular task because the geographic distribution of the Alps100K dataset, which corresponds to reality, exhibits many strong peaks at popular places (see Fig.~\ref{fig:dataset_histogram}, right). 
CNN, on the other hand, seems to generalize better on previously unseen locations and it surpasses BOW in sparsely covered regions. 
The combination of BOW and CNN achieves the best RMSE scores, and is more robust as well.

\begin{table}[t]
\centering%
  \caption{Results (overall root-mean-square error in meters)}
  \setlength{\tabcolsep}{10pt}
  \begin{tabular}{l|cc}
	\hline
	\hline
	Method       & test dataset (13148 images) & user experiment set (50 images) \\
	\hline
    Baseline   & 801.49; \textit{786.42}     & 1383.64; \textit{1154.43} \\
    Human      & -         & \textbf{879.95}\\
    CNN        & 537.11    & 709.10\\
	BOW        & 601.63    & 757.76\\
    mVocab     & 610.36    & 811.00\\
    BOW+mVocab & 564.14    & 646.89\\
    BOW+CNN    & \textbf{500.44}    & \textbf{531.05}
   \end{tabular}
  \label{tab:results}
\end{table}

The results for a subset of 50 images selected from the test dataset (used in user experiment described in Section~\ref{sec:human}) are shown in the right column of Tab.~\ref{tab:results}, which compares the performance of automatic elevation estimation to the performance of humans. Generally, all of the proposed methods achieve better scores than humans. The best RMSE is again obtained by the hybrid combination of BOW and CNN, which is on average significantly better than humans. In Fig.~\ref{fig:boxplot} we plot the predictions for each of the 50 images separately. Interestingly, BOW+CNN method exhibits a similar bias as humans; i.e., it tends to underestimate the highest elevations (images \#49, 48, 45, 41, 39, 38) and overestimate lowest elevations (\#2, 3). This tendency actually holds true for the whole test dataset, as illustrated in~Fig.~\ref{fig:cumul_graph} right. We attribute this behavior to the distribution of elevations in Alps100K dataset, which is less populated in both elevation extremes, as shown in  Fig~\ref{fig:dataset_histogram}, left.

\paragraph{CNN+EXIF information.}
Visual appearance of photographs is strongly influenced by the season, the time of the day, and camera field of view. We encode time of the day, time of the year, exposure coefficient, and camera field of view as a sparse binary vector and input it to the first fully connected layer of the CNN (fc6).
Each of the values is quantized to 16 discrete levels and encoded as 1-of-N.
%
%
The exposure coefficient ($EC$) combines relative aperture $N$, exposure time $t$, and sensor sensitivity $ISO$ into a single value that represents ``sensitivity of the photograph'' to the light in the scene. Assuming that photos are properly exposed, high $EC$ values imply low-light conditions and low $EC$ values imply bright conditions. $EC$ is calculated as
$EC = \log_2 {N^2} - \log_2 {t \cdot \frac{ISO}{100}}$.
Camera \emph{field of view} ($FOV$) indicates possible composition of photographs. As most cameras do not store their field of view explicitly in EXIF data, it needs to be computed from sensor size $S$ and focal length $f$ as $FOV = \arctan \left (  0.5 S / f \right )$.

The combination of CNN with EXIF information was evaluated on a smaller subset of images with EXIF information available (36050 training and 5314 test images respectively). On this subset, CNN+EXIF achieves RMSE=510m, compared to pure CNN with RMSE=550m on the same subset. We conclude that the EXIF data bring a small improvement in elevation predictions.

\begin{figure}[t]
	\centering
        \begin{overpic}[width=0.32\linewidth]{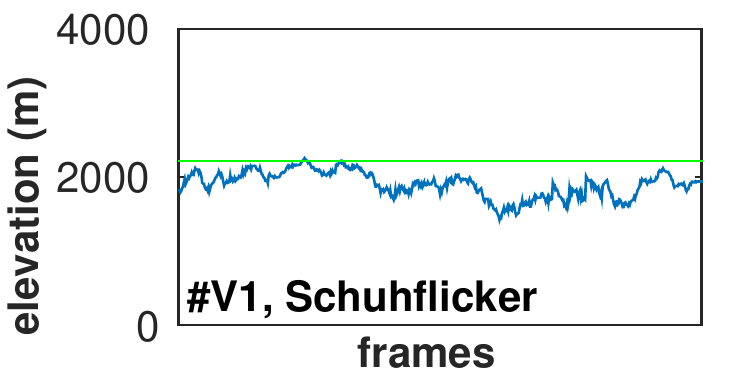}
        \end{overpic}
        \begin{overpic}[width=0.32\linewidth]{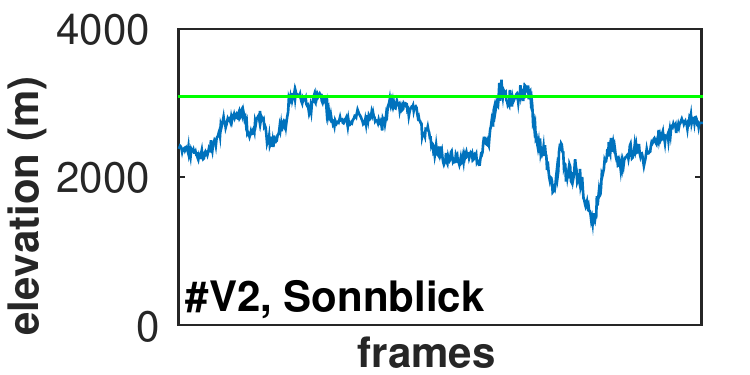}
        \end{overpic}        
        \begin{overpic}[width=0.32\linewidth]{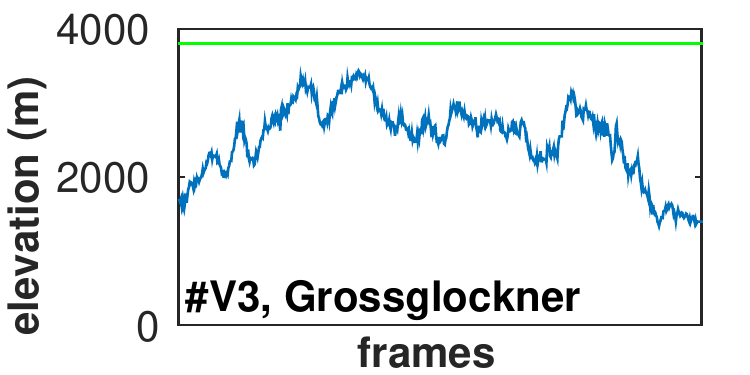}
        \end{overpic}        
	\caption{BOW+CNN elevation predictions (blue) for three video sequences (green) captured using a camera with constant elevation and varying yaw and pitch.}
	\label{fig:video_elevation}
\end{figure}

\paragraph{Video sequences.} We evaluated the BOW+CNN method using three video sequences (\#V1-3 available at the project webpage~\cite{cadik2015supplementary}) in Fig.~\ref{fig:video_elevation}.
The videos were acquired by a hand-held point-and-shoot camera from spots of constant elevation (green lines), while changing camera yaw and pitch. 
These videos represent challenging scenarios, in particular, due to the varying pitch and high camera elevations (\#V1=2215m, \#V2=3088m, \#V2=3789m). Moreover, no frame has been spatially verified by BOW in any case, and thus the prediction accuracy depends solely on CNN. Nevertheless, the proposed method achieves decent elevation predictions for \#V1 and \#V2. The third video (\#V3) illustrates the limitations of the current solution: the prediction accuracy for extremely high elevations is low, suffering from a small number of appropriate images in the training set. 

\section{Conclusions}
\label{sec:conclusions}

Multiple aspects of the camera elevation estimation task were addressed in the paper. A new benchmark dataset of elevation-annotated images Alps100K was collected. Two approaches were proposed to automatically estimate the camera elevation from a single landscape photograph. In an extensive user experiment, human performance on this task was measured. Experimental evaluation showed that the proposed methods outperform human abilities in camera elevation estimation. The best performing method first attempts to recognize specific location via BOW-based image retrieval, and in case of failure, uses CNN to estimate the elevation.
In the future, we plan to extend the dataset to different geographic areas and other climate zones.


\subsection*{Acknowledgements}
{\small This work was supported by SoMoPro II (financial contribution from the EU 7 FP People Programme Marie Curie Actions, REA 291782, and from the South Moravian Region), SGS15/155/OHK3/2T/13, and MSMT LL1303 ERC-CZ grants. 
The content of this article does not reflect the official opinion of the European Union. Responsibility for the information and views expressed therein lies entirely with the authors. 
}

\bibliography{main}
\end{document}